\documentclass{article} % For LaTeX2e
\usepackage{iclr2015,times}
\usepackage{hyperref}
\usepackage{url}

\usepackage{amssymb}
\usepackage{amsmath}
\usepackage{graphicx}
\usepackage{framed}
\usepackage{algorithm}
\usepackage{algpseudocode}

\title{Memory Networks}

\author{
Jason Weston, Sumit Chopra \& Antoine Bordes \\
Facebook AI Research \\
770 Broadway \\
New York, USA \\
\texttt{\{jase,spchopra,abordes\}@fb.com} 
}

\newcommand{\m}{{\bf m}}
\newcommand{\inp}{{x}} % instead of {\m_M}

\DeclareMathOperator*{\argmax}{arg\,max}

\iclrfinalcopy % Uncomment for camera-ready version

\iclrconference % Uncomment if submitted as conference paper instead of workshop

\begin{document}

\maketitle

\begin{abstract}
  We describe a new class of learning models called {\em memory
  networks}. Memory networks reason with inference components combined
  with a long-term memory component; they learn how to use these jointly.
  The long-term memory can be read and written to, with
  the goal of using it for prediction.
%effectively acts as a knowledge base
%that the model can learn to read and write to, and to utilize for
%e.g. prediction. , or producing a textual response.
  We investigate these models in the context of question answering
  (QA) where the long-term memory effectively acts as a (dynamic)
  knowledge base, and the output is a textual response.  We evaluate
  them on a large-scale QA task, and a smaller, but more complex, toy
  task generated from a simulated world.  In the latter, we show the
  reasoning power of such models by chaining multiple supporting
  sentences to answer questions that require understanding the intension %meaning
  of verbs.
\end{abstract}

\section{Introduction}
\label{sec:intro}

Most machine learning models lack an easy way to read and write to
part of a (potentially very large) long-term memory component, and to
combine this seamlessly with inference.  Hence, they do not take
advantage of one of the great assets of a modern day computer.  For
example, consider the task of being told a set of facts or a story,
and then having to answer questions on that subject. 
% as illustrated in the simple example of Figure~\ref{fig:story1}.
In principle
this could be achieved by a language modeler such as a recurrent
neural network (RNN) \citep{mikolov2010recurrent,hochreiter1997long}
 %or long short term memory (LSTM)
%\cite{lstm}, 
as these models are
trained to predict the next (set of) word(s) to output after having
read a stream of words.  However, their memory (encoded by hidden
states and weights) is typically too small, and is not
compartmentalized enough to accurately remember facts from the past
(knowledge is compressed into dense vectors).
 RNNs are known to
have difficulty in performing memorization, for example the simple copying task of
outputting the same input sequence they have just read \citep{zaremba2014learning}.
The situation is similar for other tasks, e.g., in the vision and audio domains a long term
memory is required to watch a movie and answer questions about it. %, something typical models lack.

In this work, we introduce a class of models called memory networks
that attempt to rectify this problem.  The central idea is to combine
the successful learning strategies developed in the machine learning
literature for inference with a memory component that can be read and
written to. The model is then trained to learn how to operate
effectively with the memory component.
%
%After introducing the general framework of memory networks in
%Section~\ref{sec:prior}, we present a first implementation in the text
We introduce the general framework in %of memory networks in
Section~\ref{sec:mem-nets}, and present a specific implementation in the text
domain for the task of question answering in Section \ref{sec:memnn1}.
We discuss related work in Section \ref{sec:related}, describe
our experiments in \ref{sec:experiments}, and
finally conclude in Section \ref{sec:conclusion}.

%The rest of the paper is as f

%Our approach was inspired by question answering systems that operate
% with a fixed knowledge base such as \cite{ecml,emnlp}. This is an
% attempt to generalize that approach to a more dynamic setting.  To
% that end, we first introduce the general class of

\section{Memory Networks}
\label{sec:mem-nets}

A memory network consists of a memory $\m$ (an array of objects\footnote{For example an array of vectors or an array of strings.}
indexed by $\m_i$) and
%has
 four (potentially learned) components %(which may or may not have shared
%parameters), they are
 $I$, $G$, $O$ and $R$ as follows:

\begin{itemize}
\item[I:] (input feature map) -- converts the incoming input to the internal feature representation.
\item[G:] (generalization) -- updates old memories given the new input. We call this generalization as there is an opportunity for the network to compress and generalize its memories at this stage for some intended future use.
\item[O:] (output feature map) --  produces a new output (in the
  feature representation space), given the new input and the current memory state.
\item[R:] (response) -- converts the output into the response format desired. For example, a textual response or an action.
\end{itemize}

%The flow of the model at prediction time is as follows. When 
%an input $x$ is observed (e.g. a sentence, an image or audio signal),
% the following steps are performed:
%The flow of the model at prediction time is as follows, given
%an input $x$  (e.g. an input character, word or sentence depending on the granularity chosen, 
%an image or an audio signal):

Given an input $x$  (e.g., an input character, word or sentence depending on the granularity chosen,  an image or an audio signal) the flow of the model is as follows:
% the following steps are performed:
\begin{enumerate}
\item Convert $x$ to an internal feature representation $I(x)$.
\item Update memories $\m_i$ given the new input: ~ $\m_i  = G(\m_i, I(x), \m), ~ \forall i$.
\item Compute output features $o$ given the new input and the memory: $o  = O(I(x), \m)$.   
\item Finally, decode output features $o$ to give the final response:  $r = R(o)$.
\end{enumerate}

This process is applied at both train and test time, if there is a distinction between such phases,
that is, memories are also stored at test time, but the model parameters of I, G, O and R are not updated.
Memory networks cover a wide class of possible implementations.  The
components $I$, $G$, $O$ and $R$ can potentially use any existing
ideas from the machine learning literature, e.g., make use of your favorite
models (SVMs, decision trees, etc.). % although neural networks are a natural choice if one
%wanted to perform end-to-end training of the components $I$, $G$, $O$ and $R$.
% use of support
%vector machines, decision trees or your favorite model
%architectures.

\paragraph{$I$ component:}
Component $I$ can make use of standard
pre-processing, e.g., parsing, coreference and entity resolution for
text inputs.  It could also encode the input into an internal feature representation,
e.g., convert from text to a sparse or dense feature vector.

\paragraph{$G$ component:}
The simplest form of $G$ is to store $I(x)$ in a ``slot'' 
in the memory:
\begin{equation}
 \m_{H(x)} = I(x),
\end{equation}
 where $H(.)$
 is a function selecting the slot. That is, $G$ updates the index $H(x)$ of $\m$, but all other parts of the memory remain untouched.
More sophisticated variants of $G$ could go back and update earlier stored memories (potentially, all memories) based on the new evidence from the current input $x$.
 If the input is at the character or word level one
could group inputs (i.e., by segmenting them into chunks) and
store each chunk in a memory slot.

If the memory is huge (e.g., consider all of Freebase or Wikipedia) one needs to organize the memories. This can be achieved with the slot choosing function $H$ just described: for example, it could be designed, or trained, to store memories by entity or topic.
Consequently, for efficiency at scale, $G$ (and $O$) need not operate on all
memories: they can operate on only a retrieved subset of candidates (only operating on memories that are on the right topic). We explore a simple variant of this in our experiments.

If the memory becomes full, a procedure for ``forgetting'' could also be implemented
by $H$ as it chooses which memory is replaced, e.g., $H$ could score
the utility of each memory, and overwrite the least useful. We have not explored this experimentally yet.

\paragraph{$O$ and $R$ components:}
The $O$ component is typically responsible for reading from memory and performing inference,
e.g., calculating what are the relevant memories to perform a good response.
The $R$ component then produces the final response given $O$.
For example in a question answering setup $O$ finds relevant memories, and then $R$ produces the
actual wording of the answer, e.g., $R$ could be an RNN that is conditioned on the output of $O$. 
Our hypothesis is that without conditioning on such memories, such an RNN will perform poorly.

%\paragraph{Efficiency for Large Memory Sizes:}

%\paragraph{Forgetting:}

\section{A MemNN Implementation For Text}
\label{sec:memnn1}

One particular instantiation of a memory network is where the
components %$I$, $G$, $O$ and $R$ 
are neural networks. We refer to
these as memory neural networks (MemNNs).  In this section we
describe a relatively simple implementation of a MemNN with textual
input and output.

\subsection{Basic model} \label{sec:basicmodel}

%An embedding-based MemNN implementation}

%We now describe a MemNN architecture developed for experimenting on
%QA. The data is assumed to arrive as a stream of textual inputs.
% Some inputs are statements that describe facts while
%others are questions about those facts to be answered by the system.

%We first describe the basic architecure, and then follow that with some extensions.
%\begin{itemize}
%\item 
%We now describe a MemNN architecture developed for QA.
In our basic architecture,
the $I$ module takes an input text.
 Let us first assume this to be a sentence: either the statement of a
 fact, or a question to be answered by the system (later we will consider word-based input sequences).
The text is stored in the next available memory slot in its original form\footnote{Technically, we will be using an embedding model to represent text, so we could store the incoming input using its learned embedding vector in memory instead. The downside of such a choice is that during learning the embedding parameters are changing, and hence the stored vectors would go stale. However, at test time (where the parameters are not changing) storing as embedding vectors could make sense, as this is faster than reading the original words and then embedding them repeatedly.},
i.e., $S(x)$ returns the next empty memory slot $N$: $\m_N = x$, $N=N+1$.  
The $G$ module is thus only used to store this new memory, so old memories are not updated.
More sophisticated models are described in subsequent sections.
%Extensions leading to more sophisticated models are described in subsequent sections.

%We now describe a MemNN architecture developed for QA.
%At each step, the $I$ module consumes an incoming input word $x$.
%Input words form sentences: either the conjunction of one or more facts, 
%or a question to be answered by the system.
%We assume such sentences, and their meaningful consituent phrases are not pre-segmented,
%and so the $I$ component decides at each step whether to segment the last set of words
%via a learned ``segmentation'' function $seg(x)$, to be described in the following.
%This is learned such that if  $seg(x) > 1$ then that sequence is represented 
 
%
%\item
%
%\item
%The $G$ module is not used, i.e. the memory is assumed to be infinite, %the memory capacity is supposed to be
%infinite and 
%and old memories are not updated.

%\item

The core of inference lies in the $O$ and $R$ modules.
The $O$ module produces output features by finding $k$ supporting
memories given $x$. We use $k$ up to $2$, but the procedure is generalizable to larger $k$. 
For $k=1$ the highest scoring supporting memory is retrieved with:
\begin{equation}
      o_1 = O_1(\inp, \m) =  \argmax\limits_{i=1,\dots,N} ~s_O(\inp, \m_i)
\label{eq:o1}
\end{equation}
where $s_{O}$ is a function that scores the match between the pair of sentences
$x$ and $\m_i$. % and $o_1\in[1;N]$.
For the case $k = 2$ we then find a second supporting memory given the first found in the previous iteration:
%In the case where $k=2$, the second supporting memory is retrieved
%using $\inp$ and the first supporting memory found in the previous iteration: 
%(via eq. (\ref{eq:o1})):
\begin{equation}
      o_2 = O_2(\inp, \m) =   \argmax\limits_{i=1,\dots,N}  ~s_O([\inp, \m_{o_1}], \m_i)
\label{eq:o2}
\end{equation}
where the candidate supporting memory $\m_i$ is now scored with respect to both the original
input and the first supporting memory, where square brackets
denote a list\footnote{As we will use a bag-of-words model where both $\inp$ and $\m_{o_1}$ are represented in the bag (but with two different dictionaries) this is equivalent to using the sum $s_O(\inp, \m_i) + s_O(\m_{o_1}, \m_i)$, however a more sophisticated modeling of the inputs (e.g., with nonlinearities) may not separate into a sum.}.
The final output $o$ is $[x, \m_{o_1}, \m_{o_2}]$, which is input to the module $R$.
%Overall, the module $O$ can be thought of as related to a sequence of  $k$ $\argmax$-pooling layers (rather than $\max$-pooling layers) that operate over the memory.

Finally, $R$ needs to produce a textual response $r$. The simplest 
response is to return $\m_{o_k}$, i.e., to output
the previously uttered sentence we retrieved. 
To perform true sentence generation, one can instead employ an
RNN. %, see e.g. recent work on machine translation \cite{rnn}.
%In our experiments we consider those cases, and also
In our experiments we %do consider those two choices, and we also tried
also consider an easy to evaluate compromise approach where
%However, for simplicity in our experiments
we limit textual responses to be a single word (out of all the words
seen by the model) by ranking them:
\begin{equation} \label{eq:R-word-score}
    r  =  {\mbox{argmax}}_{w \in W}  ~s_R([\inp, \m_{o_1}, \m_{o_2}], w)
\end{equation}
where $W$ is the set of all words in the dictionary, and $s_R$ is a
function that  scores the match. 

An example task is given in Figure~\ref{fig:story1}.
In order to answer the question
$\inp$ = ``Where is the milk now?'', the $O$ module first scores
all memories, i.e., all previously seen sentences, against $\inp$ to
retrieve the most relevant fact, $\m_{o_1}$ = ``Joe left the milk'' in
this case. Then, it would search the memory again to find the second
 relevant fact given $[\inp,\m_{o_1}]$, that is $\m_{o_2}$ = ``Joe travelled to the office'' (the last place Joe went before dropping the milk).
Finally, the $R$ module using eq. (\ref{eq:R-word-score}) 
would score words given $[\inp, \m_{o_1}, \m_{o_2}]$ to output $r$ = ``office''.

In our experiments, the scoring functions 
$s_O$ and $s_R$ have the same form, that of an embedding model:
\begin{equation}
  s(x,y) = \Phi_x(x)^\top U^\top U \Phi_y(y).
\end{equation}
where $U$ is a $n \times D$ matrix where $D$ is the number of features
and $n$ is the embedding dimension.  The role of
$\Phi_x$ and $\Phi_y$ is to map the original text to the
$D$-dimensional feature space.  The simplest feature space to choose
is a bag of words representation, we choose $D = 3|W|$ for $s_O$,
i.e., every word in the dictionary has three different representations:
one for $\Phi_y(.)$ and two for $\Phi_x(.)$ depending on whether the
words of the input arguments are from the actual input $x$ or from
the supporting memories %$\m_{o_1}$ 
so that they can be modeled
differently.\footnote{Experiments with only a single dictionary and linear embeddings performed worse (not shown).
 In order to model with only a single dictionary,
  one could consider deeper networks that transform the words
  dependent on their context. We leave this to future
  work.} Similarly, we used $D = 3|W|$ for $s_R$ as well. 
$s_O$ and $s_R$ use different weight matrices $U_O$ and $U_R$. 

\paragraph{Training}
We train in a fully supervised setting where we are given desired inputs and responses, and
the supporting sentences are labeled as such in the training data (but not in the test data, where we are given only the inputs). That is, during training we know the best choice of both
max functions in eq. (\ref{eq:o1}) and (\ref{eq:o2})\footnote{
%This information is often available in real question answer datasets, see e.g., \citep{berant2013semantic} and \citep{bordes2014question}. 
However, note that methods like RNNs and LSTMs cannot easily use this information.}.
Training is then performed with a margin ranking loss and stochastic gradient descent (SGD).
Specifically, for a given question $x$ with true response $r$ and supporting sentences
$\m_{o_1}$ and $\m_{o_2}$ (when $k=2$),  we minimize over model parameters $U_O$ and $U_R$:
\begin{eqnarray} \label{eq:train}
&  \sum\limits_{\bar{f} \neq \m_{o_1}}  \max(0, \gamma - s_O(x, \m_{o_1}) + s_O(x, \bar{f})) +   \\
&  \sum\limits_{\bar{f'} \neq \m_{o_2}} \max(0, \gamma - s_O([x, \m_{o_1}], \m_{o_2}]) + s_O([x, \m_{o_1}], \bar{f'}])) +  \label{eq:train2}  \\ 
&   \sum\limits_{\bar{r} \neq r} 
      \max(0, \gamma - s_R([x, \m_{o_1}, \m_{o_2}], r) + s_R([x, \m_{o_1}, \m_{o_2}], \bar{r}])) \label{eq:train3}
\end{eqnarray}
where $\bar{f}$, $\bar{f'}$ and $\bar{r}$ are all other choices than the correct labels,
and $\gamma$ is the margin.
At every step of SGD we sample $\bar{f}, \bar{f'}, \bar{r}$ rather than compute the whole 
sum for each training example, following e.g., \cite{wsabie}.

In the case of employing an RNN for the $R$ component of our MemNN 
(instead of using a single word response as above)
 we replace the last term 
with the standard log likelihood  used in a language modeling task,
where the RNN is fed the sequence $[x, o_1, o_2, r]$. At test time we  output its prediction $r$ given $[x, o_1, o_2]$.
In contrast the absolute simplest model, that of using $k=1$ and outputting the located memory 
$\m_{o_1}$ as response $r$, would only use the first term to train.

%language modeling task
%greedy word sequence log likelihood optimization of an RNN conditioned on $[x, o_1, o_2]$, see e.g. \cite{which?}.
%We train these models using stochastic gradient descent given 
%supervised data of the supporting 
% similarly to the {\sc Wsabie} model training procedure

In the following subsections we consider some extensions of our basic model.

\begin{figure}
\caption{Example ``story'' statements, questions and answers generated by a simple simulation.
Answering the question about the location of the milk requires comprehension of the actions ``picked up'' and ``left''. The questions also require comprehension of the time elements of the story, e.g., to answer ``where was Joe before the office?''. % requires comprehension of the time elements of the story.
\label{fig:story1}}
\begin{small}
\begin{framed}
Joe went to the kitchen. 
Fred went to the kitchen.
Joe picked up the milk.\\
Joe travelled to the office.
Joe left the milk.
 Joe went to the bathroom.\\
Where is the milk now? \textcolor{red}{A: office}\\
Where is Joe? \textcolor{red}{A: bathroom} \\
Where was Joe before the office? \textcolor{red}{A: kitchen}
\end{framed}
\end{small}
\end{figure}

\subsection{Word Sequences as Input} \label{sec:word-seq}

If input is at the word rather than sentence level, that is words arrive in a stream
 (as is often done, e.g., with RNNs) 
and not already segmented as statements and questions, 
we need to modify the approach we have so far described.
%we need to modify the approach we have so far described.
%In such a setting sentences, and their meaningful consituent phrases, are not already
%pre-segmented as statements and questions.
We hence add a ``segmentation'' function, to be learned, which takes as input 
the last sequence of words 
that have so far not been segmented and looks for breakpoints.
When the segmenter fires (indicates the current sequence is a
 segment) we write that sequence to memory, and can then proceed as before.
The segmenter is modeled similarly to our other components, as an embedding model
of the form:
\begin{equation}
  seg(c) = W_{seg}^\top U_{S} \Phi_{seg}(c) 
%         seg(c) = U_{S} \Phi_{seg}(c) 
\end{equation}
where $ W_{seg}$ is a vector (effectively the parameters of a linear classifier in embedding space),
and $c$ is the sequence of input words represented as bag of words using a separate dictionary.
If $seg(c) > \gamma$, where $\gamma$ is the margin, then this sequence is recognised as a segment.
In this way, our MemNN has a learning component in its write operation.
We consider this segmenter a first proof of concept: of course, one could design something much
more sophisticated.
Further details on the training mechanism are given in Appendix \ref{app:wordseq-training}. 

%then it seems natural to group such words into meaningful segments for storage in memory.
%For example, if input sentences contain a conjunction of more than one fact, segementation
%into multiple meaningful consituent phrases could be beneficial.
%To implement this, we consider at each step considering the sequence of 
%At each step, the $I$ module consumes an incoming input word $x$.
%Input words form sentences: either the conjunction of one or more facts, 
%or a question to be answered by the system.
%This is learned such that if  $seg(x) > 1$ then that sequence is represented 

\subsection{Efficient Memory via Hashing} \label{sec:memhash}

If the set of stored memories is very large it is prohibitively expensive to score all of them as in 
equations (\ref{eq:o1}) and (\ref{eq:o2}).
Instead we explore hashing tricks to speed up lookup: hash the input $I(x)$ into one or more buckets and then only score memories $\m_i$ that are in the same buckets.
We investigated two ways of doing hashing: (i) via hashing words; and (ii) via clustering word embeddings. For (i) we construct as many buckets as there are words in the dictionary, then for a given sentence we hash it into all the buckets corresponding to its words. The problem with (i) is that a memory $\m_i$ will only be considered if it shares at least one word with the input $I(x)$.
Method (ii) tries to solve this by clustering instead. After training the embedding matrix $U_O$, 
 we run $K$-means to cluster word vectors $(U_O)_i$, thus giving $K$ buckets.
 We then hash a given sentence into all the buckets that its individual words fall into.
As word vectors tend to be close to their synonyms, they cluster together and we thus also will score those similar memories as well. Exact word matches between input and
 memory will still be scored by definition. Choosing $K$ controls the speed-accuracy trade-off.

\subsection{Modeling Write Time}\label{sec:writetime}

We can extend our model to take into account {\em when} a memory slot
was written to.  This is not important when answering questions about
fixed facts (``What is the capital of France?'')  but is important
when answering questions about a story, see
e.g., Figure \ref{fig:story1}.  One obvious way to implement this is to
add extra features to the representations $\Phi_x$ and $\Phi_y$ 
that encode the index $j$ of a given memory $\m_j$, assuming that $j$
follows write time (i.e., no memory slot rewriting).  
However, that
requires dealing with absolute rather than relative time.  We had more
success empirically with the following procedure: instead of scoring
input, candidate pairs with $s$ as above, learn a function on triples  $s_{O_t}(x,y,y')$:% defined as:
%that scores triples:
\begin{equation}
  s_{O_t}(x,y,y') = \Phi_x(x)^\top {U_{O_t}}^\top {U_{O_t}} \Big(\Phi_y(y) -\Phi_y(y') + \Phi_t(x,y,y')\Big).
\end{equation}
$\Phi_t(x,y,y')$ uses three new features which take on the value 0 or
1: whether $x$ is older than $y$, $x$ is older than $y'$, and $y$
older than $y'$.  (That is, we extended the dimensionality of all the $\Phi$ embeddings by 3, and set these three dimensions to zero when not used.)
% In a $k=2$ model the first two features use the time stamp of  $\m_{o_1}$ in the second loop.
Now, if $s_{O_t}(x,y,y') > 0$ the model prefers $y$ over
$y'$, and if $s_{O_t}(x,y,y')<0$ it prefers $y'$.  The $\mbox{argmax}$ of
eq. (\ref{eq:o1}) and (\ref{eq:o2}) are replaced by a loop over
memories $i=1,\dots,N$, keeping the winning memory ($y$ or $y'$) at each step,
and always comparing the current winner to the next memory
$\m_i$. This procedure is equivalent to the $\mbox{argmax}$
before if the time features are removed.
More details are given in Appendix \ref{app:time-training}.

\subsection{Modeling Previously Unseen Words}
\label{sec:new-words}

Even for humans who have read a lot of text, new words are continuously
introduced.  For example, the first time the word ``Boromir''
appears in Lord of The Rings {\citep{tolkien2012fellowship}}.
%
%{\em `Here,' said Elrond, turning to Gandalf, `is Boromir, a man from the South. He arrived in the grey morning, and seeks for counsel\dots”}
%seen only one example.
How should a machine learning model deal with this? Ideally it should work having  seen only one example.
%Ideally a machine learning model should work having  seen only one example.
%react and work instantaneously right. 
A possible way would be to use a
language model: given the neighboring words, predict what the word
should be, and assume the new word is similar to that.
Our proposed
approach
%can be thought of like that, 
takes this idea, but incorporates it into our networks $s_O$ and $s_R$,
rather than as a separate step.

Concretely, for each word we see, we store a bag of words it has
co-occurred with, one bag for the left context, and one for the
right. Any unknown word can be represented with such features.  Hence,
we increase our feature representation $D$ from $3|W|$ to $5|W|$ to
model these contexts ($|W|$ features for each bag).  
Our model learns
to deal with new words during training using a kind of ``dropout''
%technique: sometimes we pretend we have not seen a word before, and hence
technique: $d\%$ of the time we pretend we have not seen a word before, and hence
do not have a $n$-dimensional embedding for that word, and represent it with 
the context instead.
% JW: I put this back as yours doesn't seem right to me.
%technique to mimic occurrences of unknown words: sometimes era word is
%pretended to be unknown and hence do not have a
%$d$-dimensional embedding in $U$, and is represented by its complex
%instead. {\bf AB: context is used for known words as well or not?} JW: Nope.

\subsection{Exact Matches and Unseen Words}
\label{sec:matchf}

Embedding models cannot efficiently use exact word matches due to the low
dimensionality $n$. % (see e.g. \cite{bai}). 
One solution is to score a pair $x, y$ with
\begin{equation}
 \Phi_x(x)^\top U^\top U \Phi_y(y) +  \lambda  \Phi_x(x)^\top  \Phi_y(y)
\end{equation}
instead. That is, add the ``bag of words'' matching score to the
learned embedding score (with a mixing parameter $\lambda$).  Another,
related way, that we propose is to stay in the $n$-dimensional
embedding space, but to extend the feature representation $D$ with
{\em matching features}, e.g., one per word. A matching feature
indicates if a word occurs in both $x$ and $y$.  That is, we score with
$\Phi_x(x)^\top U^\top U \Phi_y(y, x)$ where $\Phi_y$ is actually
built conditionally on $x$: if some of the words in $y$ match the words
in $x$ we set those matching features to 1.  Unseen words can be
modeled similarly by using matching features on their
context words. %instead.
This then gives a feature space of $D=8|W|$.
% {\bf AB: this leads to $D=6|W|$ or even a $D=8|W|$ to encode match features?} JW: Yes, at least 2 more bags.

%\subsection{Training}
%We train these models using stochastic gradient descent given 
%supervised data of the supporting 
% similarly to the {\sc Wsabie} model training procedure

\section{Related Work} \label{sec:related}

Classical QA methods use a set of documents as a kind of memory, and information retrieval methods
to find answers, see e.g., \citep{kolomiyets2011survey}  and references therein.
More recent methods try instead to create a graph of facts -- a knowledge base (KB) -- as their
memory, and map questions to logical queries \citep{berant2013semantic,berant2014modeling}.
Neural network and embedding 
approaches have also been recently explored \citep{bordes2014question,iyyer2014neural,export:214353}.
Compared to recent knowledge base approaches, memory networks differ in that they do not apply a two-stage strategy:
(i) apply information extraction principles first to build the KB; followed by (ii) inference over the KB.
Instead, extraction of useful information to answer a question is performed on-the-fly 
over the memory which can be stored as raw text, as well as other choices such as embedding vectors.
This is potentially less brittle as the first stage of building the KB may have already thrown away the relevant
part of the original data.

Classical neural network memory models such as associative memory networks
aim to provide content-addressable memory, i.e., given
a key vector to output a value vector, see e.g., \citet{haykin1994neural} and references therein.
 Typically this type of memory is distributed across
the whole network of weights of the model  rather than being compartmentalized into memory locations.
Memory-based learning such as nearest neighbor, on the other hand, does seek to store all (typically 
labeled) examples in compartments in memory, but only uses them for finding closest labels.
 Memory networks combine compartmentalized  memory 
with neural network modules that can learn how to (potentially successively) 
read and write to that memory, e.g., to perform reasoning they can iteratively read
salient facts from the memory.

However, there are some notable models that have attempted to include memory read and write operations from the 90s. In particular \citep{das1992learning}
 designed  differentiable push and pop actions called a neural network pushdown automaton. The work of \citet{schmidhuber1992learning} incorporated the concept of two neural networks where one has very fast changing weights which can potentially be used as memory. \citet{schmidhuber1993self} proposed to allow a network to modify its own weights ``self-referentially'' which can also be seen as a kind of memory addressing. Finally two other relevant works
are the DISCERN model of script processing and memory 
\citep{miikkulainen1990discern} and the NARX recurrent networks 
for modeling long term dependencies \citep{lin1996learning}.

Our work was submitted to arxiv just before the Neural Turing Machine work of 
\citet{graves2014neural}, which is one of the most relevant related methods. 
Their method also proposes to perform (sequence) prediction using a ``large, addressable memory'' which can be read and written to. In their experiments, the memory size was limited
to 128 locations, whereas we consider much larger storage (up to 14M sentences).
The experimental setups are notably quite different also: whereas we focus on language and reasoning
tasks, their paper focuses on problems of sorting, copying and recall. On the one hand their problems require considerably more complex models than the memory network described in Section \ref{sec:memnn1}. %  (which would not be capable of sorting, for instance). 
On the other hand, their problems have known algorithmic solutions, whereas (non-toy) language problems do not.

There are other recent related works. RNNSearch \citep{bahdanau2014neural} is 
a method of machine translation that uses a learned alignment mechanism over the input sentence representation while predicting an output in order to overcome poor performance on long sentences.
The work of \citep{graves2013generating} performs %also use alignment over memory for
 handwriting recognition by dynamically determining  ``an alignment between the text and the pen locations'' so that ``it learns to decide which character to write next''. 
 One can view these as particular variants of memory networks where in that case the memory only extends back a single sentence or character sequence.

\if 0
 A/ the original QA papers that do Information Retrieval within a
collection of documents to look for an answer? This is the same as MemNNs
with Memories = documents, O looking for the correct document and R
looking for the answer in the document. See the schema on p10 of
https://lirias.kuleuven.be/bitstream/123456789/313539/1/KolomiyetsMoensIS20
11.pdf . We could almost label some boxes with I, O and R. (G is less
evident). Perhaps we can see anything like this, but considering our
application (QA), this is highly relevant I think. For differences are:
    - the time features (if we use them)
    - the G component if we organize the memory during training or
a-posterori
    - the recursively (k>1) that seems complicated in the standard setting
(but possible if you look at the schema by using the ³Dialog Question²
path)
  B/ the papers that try to perform QA on stories while first ³reading²
the story and creating a KB out of it. The best example is the last paper
of Berant at EMNLP14 (http://www.aclweb.org/anthology/D14-1159.pdf).
Instead of organizing the memory as a stack of raw facts, they create a
graph of events, which is later queried. I think this also highly relevant.
\fi

\section{Experiments}
\label{sec:experiments}

\subsection{Large-scale QA}
\label{sec:fader}

We perform experiments on the QA dataset introduced in \citet{paralex}.
It consists of 14M statements, stored as (subject, relation, object)
triples, which are stored as memories in the MemNN model. The triples
are {\em REVERB} extractions mined from the ClueWeb09 corpus and cover
diverse topics such as {\em(milne, authored, winnie-the-pooh)} and
{\em (sheep, be-afraid-of, wolf)}.  Following
\citet{paralex} and \citet{bordes2014open},
 training combines pseudo-labeled QA
pairs {made of a question and an associated triple}, and 35M pairs of
paraphrased questions from WikiAnswers like {\em ``Who wrote the
  Winnie the Pooh books?''} and {\em ``Who is poohs creator?''}.

{We performed experiments in the framework of re-ranking the top
returned candidate answers by several systems measuring F1 score over the test set,
following \citet{bordes2014open}. These answers have been annotated as right or wrong
by humans, whereas other answers are ignored at test time as we do not know their label.
 We used a MemNN model of Section
\ref{sec:memnn1} with a $k=1$ supporting memory, which ends up being
similar to the approach of \citet{bordes2014open}.\footnote{We use a
  larger 128 dimension for embeddings, and no fine tuning, hence the
  result of MemNN slightly differs from those reported in
  \citet{bordes2014open}.}}  We also tried adding the bag of words
features of Section \ref{sec:matchf} as well.  Time and unseen word
modeling were not used. 
 Results are given in Table 
\ref{table:fader-qa}. %, compare to existing approaches when re-ranking
%the top returned candidate answers over the test set, following
%\cite{bordes2014open}.
%
{The results %, along with the previous ones of \cite{bordes2014open},
show that MemNNs are a viable approach for large scale QA in terms of performance.
}
However, lookup is linear in the size of the memory, which with 14M facts is slow.
We therefore implemented the memory hashing techniques of Section \ref{sec:memhash}
using both hashing of words and clustered embeddings. For the latter we tried $K=1000$ clusters.
The results given in Table \ref{table:memhash} show that one can get significant speedups ($\sim$80x)
while maintaining similar performance using the cluster-based hash. The string hash on the other hand loses performance (whilst being a lot faster) because answers which share no words are now no longer matched.

\begin{table}
\caption{Results on the large-scale QA task of \citep{paralex}.
\label{table:fader-qa}
}
\begin{center}
\begin{tabular}{|l|c|c|}
\hline
Method & F1 \\
\hline
 \citep{paralex} %{\sc Paralex} 
                                         & 0.54 \\
 \citep{bordes2014open} %Embeddings            
                                         & 0.73  \\
%\cite{bordes2014open} Embeddings + Fine Tuning & 0 & 0 \\
MemNN (embedding only)       & 0.72  \\
MemNN (with BoW features)    & 0.82 \\
\hline
\end{tabular}
\end{center}
\end{table}
%\vspace{-0.7cm}
\begin{table}
\caption{Memory hashing results on the large-scale QA task of \citep{paralex}.
\label{table:memhash}
}
\begin{center}
\begin{tabular}{|l|c|c|c|}
\hline
Method & Embedding F1 & Embedding + BoW F1 & Candidates (speedup) \\
\hline
MemNN (no hashing)      & 0.72 & 0.82   & 14M ~~ (0x)  \\
MemNN (word hash)       & 0.63 & 0.68   & 13k  (1000x)   \\
MemNN (cluster hash)    & 0.71 & 0.80   & 177k ~ (80x)   \\
\hline
\end{tabular}
\end{center}
\end{table}

\subsection{Simulated World QA}
\label{sec:simulation}

%\paragraph{Experimental Setup}

Similar to the approach of \citet{bordes2010towards} we also built a
simple simulation of 4 characters, 3 objects and 5 rooms -- with
characters moving around, picking up and dropping objects.  The
actions are transcribed into text using a simple automated grammar,
and labeled questions are generated in a similar way.  This gives a QA
task on simple ``stories'' such as in Figure \ref{fig:story1}. 
The overall difficulty of the task is that multiple
statements have to be used to do inference when asking where an object is, 
e.g. to answer where is the
milk in Figure \ref{fig:story1} one has to understand the meaning of the
actions ``picked up'' and ``left'' and the influence of their relative order.
 We generated 7k statements and 3k questions from the simulator for
training\footnote{Learning curves with different numbers of training examples are given in 
Appendix \ref{app:learning-curves}.}, and an identical number for testing and compare MemNNs to
RNNs and LSTMs %\cite{hochreiter1997long} 
(long short term memory RNNs \citep{hochreiter1997long})   on this task.  
To test with  sequences of words as input (Section \ref{sec:word-seq})
the statements are joined together again with a simple grammar\footnote{We also tried the same kind of experiments with sentence-level rather than word-sequence input, without joining sentences, giving results with similar overall conclusions, see Appendix \ref{app:sentence-level}.}, 
to produce sentences that may contain multiple statements, see e.g.,
 Figure \ref{fig:rnn-story}.

{We control the complexity of the task by setting a limit on the
  number of time steps in the past the entity we ask the question about was last mentioned.
We try two experiments: using a limit of 1, and of 5, i.e., if the limit is 5 then we pick
a random sentence between 1-5 time steps in the past. {If
 this chosen sentence only mentions an actor, e.g., ``Bill is in the kitchen'' then
we generate the question ``where is Bill?'' or ``where was Bill before the kitchen?''.
If the sentence mentions an object, e.g.,
``Bill dropped the football'' then we ask the question ``where is the football?''.
For the answers we consider two options: (i) single word answers; and (ii) a simple grammar for 
generating true answers in sentence form, e.g.,  ``kitchen'' for (i) and ``He is in the kitchen I believe'' (and other variants) for (ii).
More details on the dataset generation are given in Appendix \ref{app:data-gen}.
Note that in the object case the supporting statements necessary to deduce the answer 
 may not lie in the last 5 sentences, e.g., in this example
the answer depends on other sentences to find out where Bill actually was when he dropped the football.
In fact, in the dataset
we generated necessary supporting statements
 can be up to 65 sentences before (but are usually closer).}
%It is possible to find those two supporting sentences 
%That is, those questions involve a two-step inference process, by 
%first looking at who was carrying an object last, and then
% searching                                   
%    for where this carrier was located.
For that reason, we also conducted two further types of experiments: where we only ask
questions about actors (easier) and about actors and objects (harder).
We also consider the actor-based questions without the ``before'' questions for the simplest possible task
(i.e. ``where is Bill?'' but not ``where was Bill before the kitchen?'' questions).

\begin{table}[t]
\caption{Test accuracy on the simulation QA task.
\label{table:simQA}
}
\begin{center}
\begin{tabular}{|l||c|c|c||c|c|}
\hline
        &    \multicolumn{3}{c||}{Difficulty 1} &  \multicolumn{2}{|c|}{Difficulty 5}  \\
\hline
Method &   actor w/o before & actor & actor+object &  actor & actor+object \\
\hline
% Difficulty 5:
RNN                    & 100\% & 60.9\% & 27.9\% & 23.8\% & 17.8\% \\
LSTM                   & 100\% & 64.8\% & 49.1\% & 35.2\% & 29.0\%\\
\hline
MemNN $k=1$           & 97.8\%  &  31.0\% & 24.0\%  &  21.9\% & 18.5\% \\
MemNN $k=1$ {(+time)} & 99.9\%  &  60.2\% & 42.5\%    &  60.8\% & 44.4\% \\
MemNN $k=2$ {(+time)} & 100\%    &  100\% & 100\%  &  100\% & 99.9\% \\
% Difficulty 1:
%RNN                   &  0\% & 42\%   \\
%MemNN $k=1$           &  10\% & 81\% \\
%MemNN $k=1$ {(+time)} &  0\% & 27\% \\
%MemNN $k=2$ {(+time)} &  0\% & 0.05\% \\
\hline
\end{tabular}
\end{center}
\end{table}
\begin{figure}[h!]
\caption{Sample test set predictions (in red) for the simulation in the setting 
of word-based input and 
where answers are sentences and an LSTM is used as the $R$ component of the MemNN.
\label{fig:rnn-story}}
\begin{small}
\begin{framed}
%Fred went to the office, after that Joe travelled to the bathroom, next Joe left the apple there, later Fred travelled to the kitchen.  \\
%Where is Fred?  \textcolor{red}{I believe he is in the kitchen} \\
%Where is Joe now?  \textcolor{red}{I think he is in the bathroom} \\
Joe went to the garden then Fred picked up the milk; Joe moved to the bathroom and Fred dropped the milk, and then Dan moved to the living\_room.  \\
Where is Dan?  \textcolor{red}{A: living room I believe}\\
Where is Joe? \textcolor{red}{A: the bathroom}\\
Fred moved to the bedroom and Joe went to the kitchen then Joe took the milk there and Dan journeyed to the bedroom; Joe discarded the milk. \\
Where is the milk now ? \textcolor{red}{A: the milk is in the kitchen}\\
Where is Dan now? \textcolor{red}{A: I think he is in the bedroom}\\
Joe took the milk there, after that Mike travelled to the office, then 
Joe went to the living\_room, next Dan went back to the kitchen and  
Joe travelled to the office. \\
Where is Joe now?  \textcolor{red}{A: I think Joe is in the office}
\end{framed}
\end{small}
\end{figure}

%
%The difficulty of the task is that multiple
%sentences have to be used to do inference, e.g. to answer where is the
%milk in Figure \ref{fig:story1} one has to understand the meaning of the
%actions ``picked up'' and ``left'' and the influence of their relative order.
%The supporting sentences are labeled as such in the training data for MemNN training (i.e., this is a 
%fully supervised setting). That is, during training we know the best choice of max functions
%in eq. (\ref{eq:o1}) and (\ref{eq:o2}).

For the baseline RNN and LSTM systems
%, we use a standard Elman network 
we perform language modeling 
 with backpropagation through time \citep{mikolov2010recurrent}, 
but where we backprop only on answer words\footnote{We tried using standard language modeling on the questions as well, with slightly worse results.}.
We optimized the hyperparameters: size of the hidden layer, bptt steps,  and learning rate for each dataset. 
For MemNNs we fixed the embedding dimension to 100, learning rate to 0.01 and margin $\gamma$
 to 0.1 and 10 epochs of training in all experiments.

\paragraph{Results}

The results for the single word answer setting (i) are given in Table \ref{table:simQA}.
% {The latter is a harder
%setting that actually involves to do a two-step inference process, by
%first looking at who was carrying an object last, and then searching
%for where this carrier was located.
%
For the  actor-only tasks, RNN and LSTMs
solve the simpler difficulty level 1 task {\em without} before questions (``w/o before''),
 but  perform worse {\em with}
 before questions, and even worse on the difficulty 5 tasks.
This demonstrates that the poor performance of the RNN  is due to its failure to encode
long(er)-term memory. This would likely deteriorate even further with higher difficulty levels (distances).
LSTMs are however better than RNNs, as expected, as they are designed with a more sophisticated memory model, but still have trouble remembering sentences too far in the past.
%
%
%
%MemNNs without time features also perform badly. %, as they have no notion of time relationship between utterances.
%but still outperform RNNs on the difficulty 5 actor only task (although they do not completely
%solve the difficulty 1 task).
%
%
MemNNs do not have this memory limitation and its mistakes are
instead due to incorrect usage of its memory, when the wrong statement
is picked by $s_O$. Time features are necessary for good performance on before questions or difficulty $>$ 1 (i.e., when the answer is not in the last statement), 
otherwise %That is because a MemNN without write time features
$s_O$  can pick a statement about a person's whereabouts but they have since moved. 
%Using the time features allows to fix this
%efficiently.
 Finally, results on the harder actor+object task indicate that
MemNN also successfully perform 2-stage inference using $k=2$, 
whereas MemNNs without such inference (with $k=1$) and RNNs and LSTMs fail.}

We also tested MemNNs in the multi-word answer setting (ii) with similar results, whereby MemNNs outperform RNNs and LSTMs,
 which are detailed in Appendix \ref{app:multiword}. Example test prediction output demonstrating
the model in that setting is given in Figure \ref{fig:rnn-story}.

\subsubsection{QA with Previously Unseen Words}
We then tested the ability of MemNNs to deal with previously unseen words at test time using the
unseen word modeling approach
of Sections \ref{sec:new-words} and \ref{sec:matchf}.
We trained the MemNN on the
same simulated dataset as before and test on  the story given in
Figure \ref{fig:new-words}.
This story is generated using similar structures as in the simulation data,
except that the nouns are unknowns to the system at training time.
Despite never seeing any of the Lord of The Rings specific words before
(e.g., Bilbo, Frodo, Sauron, Gollum, Shire and  Mount-Doom),
 MemNNs are able to correctly answer the
questions. 

{MemNNs can discover simple linguistic patterns based on verbal
  forms such as (X, dropped, Y), (X, took, Y) or (X, journeyed to, Y)
  and can successfully generalize the meaning of their instantiations
  using unknown words to perform 2-stage inference.
Without the unseen word modeling described in Section \ref{sec:new-words},
they completely fail on this task.

\begin{figure}
\caption{An example story with questions correctly answered by a MemNN. The MemNN was trained on the simulation described in Section \ref{sec:simulation} and had never seen many of these  words  before, e.g., Bilbo, Frodo and Gollum.
\label{fig:new-words}
} %It relies on the new word modeling method of section \ref{sec:new-words}.
\begin{small}
\begin{framed}
Bilbo travelled to the cave. 
Gollum dropped the ring there. 
Bilbo took the ring. \\
Bilbo went back to the Shire. 
Bilbo left the ring there. 
Frodo got the ring. \\
Frodo journeyed to Mount-Doom. 
Frodo dropped the ring there. 
Sauron died. \\
Frodo went back to the Shire. 
Bilbo travelled to the Grey-havens. 
The End.\\
Where is the ring?  \textcolor{red}{A: Mount-Doom}\\
Where is Bilbo now? \textcolor{red}{A: Grey-havens}\\
Where is Frodo now? \textcolor{red}{A: Shire}
\end{framed}
\end{small}
%\end{figure}
%\begin{figure}
\caption{An example dialogue with a MemNN system trained on both the simulation data and the large-scale QA data. The system is able to (attempt to) answer questions about general world knowledge and about specific story-based statements in the dialogue. 
\label{fig:ensemble_model}
}
\begin{small}
\begin{framed}
Fred went to the kitchen.
Fred picked up the milk.
Fred travelled to the office.\\
Where is the milk ?  ~\textcolor{red}{A: office}\\
Where does milk come from ?   ~\textcolor{red}{A: milk come from cow}\\               
What is a cow a type of ?     ~\textcolor{red}{A: cow be female of cattle}\\
Where are cattle found ?      ~\textcolor{red}{A: cattle farm become widespread in brazil}\\
What does milk taste like ?   ~\textcolor{red}{A: milk taste like milk}\\
What does milk go well with ? ~\textcolor{red}{A: milk go with coffee}\\
Where was Fred before the office ? ~\textcolor{red}{A: kitchen}
%What is a kitchen for ? ~\textcolor{red}{A: kitchen be one for 4person}\\
%What is cooked in a kitchen ? ~\textcolor{red}{A: food cook in kitchen}\\
\end{framed}
\end{small}
\end{figure}

\subsection{Combining simulated data and large-scale QA}

Combining simulated world learning with real-world data might be one way to show the power
and generality of the models we design.
We implemented a naive setup towards that goal: we took the two models from Sections 
\ref{sec:fader} and \ref{sec:simulation},  trained on
large-scale QA and simulated data respectively, and built an ensemble of the two.
We present the input to both systems and then for each question simply output the response 
of the two choices with the highest score.
This allows us to perform simple dialogues with our combined MemNN system. The system
is then capable of answering both general knowledge questions and specific statements relating
to the previous dialogue. An example dialogue trace is given in Fig. \ref{fig:ensemble_model}.
Some answers appear fine, whereas others are nonsensical.
Future work should combine these models more effectively, for example
 by multitasking directly the tasks with a single model.

\section{Conclusions and Future Work}
\label{sec:conclusion}

In this paper we introduced a powerful class of models, memory networks,
and showed one instantiation for QA.
Future work should develop MemNNs for text further, evaluating them on harder QA 
and open-domain machine comprehension tasks \citep{richardson2013mctest}. For example,
large scale QA tasks that require multi-hop inference such as WebQuestions should also 
be tried \cite{berant2013semantic}.
More complex simulation data could also be constructed in order to bridge that gap,
e.g., requiring coreference, involving more verbs and nouns, sentences with more structure 
and requiring more temporal and causal understanding.
%However, we believe there are many more uses of MemNNs not explored
%here.
More sophisticated architectures should also be explored in order to deal with these tasks,
e.g., using more sophisticated memory management via $G$ and more sophisticated sentence representations. Weakly supervised settings are also very important, and should be explored, as many datasets only
have supervision in the form of question answer pairs, and not supporting facts as well as we
 used here.
%They 
%
Finally, we believe this class of models is much richer than the one specific variant we detail here,
and that we have currently only explored one specific variant of memory networks.
%For example we see \citet{bahdanau2014neural} as a particular variant of MemNNs for machine translation, where in that case the memory only extends back a single sentence.
Memory networks should be applied to other text tasks, and
other domains, such as vision, as well. 

%Make it harder/add more stuff, e.g. “he went..”, “Frodo and Sam”, etc.!!!
%MemNNs that reason with more than 2 supporting memories.
%Weakly supervised? Curriculum probably necessary.
%Ask questions? Say statements? Perform actions?

\subsubsection*{Acknowledgments}

We thank Tomas Mikolov for useful discussions.

\bibliography{refs}
\bibliographystyle{iclr2015}

\newpage
\appendix
\section{Simulation Data Generation}\label{app:data-gen}

%Our goal is to train a learner that begins 
% with little or no knowledge in an environment where 
%language is used between actors to fulfill tasks.
%We believe
%This is most easily tested in a simulated environment at this early stage.

\paragraph{Aim}
We have built a simple simulation which behaves much like a classic text adventure game.
The idea  is that generating text within this simulation allows us to ground the language used.

Some comments about our intent:
\begin{itemize}
\item Firstly, 
while this currently only encompasses a very small part of the kind of language and understanding
we want a model to learn to move 
towards full language understanding, we believe it is a prerequisite 
that models should perform well on this kind of task for  them to work on real-world
environments.
% before being tried on real world tasks.
\item Secondly, our aim is to make this simulation more complex and to release improved
 versions over time. Hopefully it can then scale up to evaluate more and more useful properties.
%test various properties we expect good language understanding models to possess.
\end{itemize}

Currently, tasks within the simulation are restricted to question answering tasks about the location of people and objects. However, we envisage other tasks should be possible, including asking the learner to perform actions within the simulation (``Please pick up the milk'', ``Please find John and give him the milk'') and asking the learner to describe actions (''What did John just do?''). % after he John has picked up the milk). 

\paragraph{Actions}
The underlying actions in the simulation consist of the following:

{\em go $<$location$>$}, ~~~
{\em get $<$object$>$}, ~~~
{\em get $<$object1$>$ from $<$object2$>$}, \\
{\em put $<$object1$>$ in/on $<$object2$>$}, ~~~
{\em give $<$object$>$ to $<$actor$>$}, \\ %~~
{\em drop $<$object$>$}, ~~~
{\em look}, ~~~
{\em inventory}, ~~~
{\em examine $<$object$>$}.

\if 0
\begin{itemize}
\item go $<$location$>$
\item get $<$object$>$
\item get $<$object1$>$ from $<$object2$>$
\item put $<$object1$>$ in/on $<$object2$>$
\item give $<$object$>$ to $<$actor$>$
\item drop $<$object$>$
\item look
\item inventory
\item examine $<$object$>$
\end{itemize}
\fi

There are a set of constraints on those actions. For example an actor cannot get something that
they or someone else already has, they cannot go to a place they are already at, cannot drop
something they do not already have, and so on.

\paragraph{Executing Actions and Asking Questions}
Using the underlying actions and their constraints,
there is then a (hand-built) model that defines how
actors act. Currently this is very simple: they try to make a random valid action,
at the moment restricted to go {\em or} go, get and drop depending on the which of two types of experiments we are running: (i) actor; or (ii) actor + object.

If we write these actions down in text form this gives us a very simple ``story''
which is executable by the simulation, e.g., 
{\em { joe go kitchen}; {fred go kitchen}; {joe get milk}; {joe go office}; {joe drop milk};
{joe go bathroom}}. This example corresponds to the story given in Figure \ref{fig:story1}.
The system can then ask questions about the state of the simulation e.g.,
{\em where milk?}, {\em where  joe?}, {\em where joe before office?}
It is easy to calculate the true answers for these questions as we have access to
the underlying world.
What  remains is to convert both the statements and the questions to look
 more like natural language.

\paragraph{Simple Grammar For Generating Language}
In order to produce more natural looking text with lexical variety we built a simple
automated grammar. Each verb is assigned a set of synonyms,
e.g., the simulation command {\em get} is replaced with  either
{\em picked up}, {\em got}, {\em grabbed} or {\em took}, and {\em drop} is replace with 
either {\em dropped}, {\em left}, {\em discarded}  or {\em put down}.
Similarly, each object and actor can have a set of replacement synonyms as well,
although currently there is no ambiguity there in our experiments, we simply add articles or not.
We do add lexical variation to questions, e.g., ``Where is John ?'' or
``Where is John now ?''.

\paragraph{Joining Statements}
Finally, for the word sequence training setting, we
join the statements above into compound sentences.
To do this we simply take the set of statements and then join them randomly  with one of 
the following:
 ``{\em .}'', ``{\em and}'', ``{\em then}'', ``{\em , then}'', ``{\em ;}'',
``{\em, later}'',  ``{\em, after that}'', ``{\em, and then}'', or ``{, next}''.
Example output can be seen in Figure \ref{fig:rnn-story}.

\paragraph{Issues}
%Here, we briefly state things the simulation does not model that we would like it to do in the future.
There are a great many aspects of language not yet modeled. For example, currently coreference is not modeled (e.g., ``He picked up the milk'') and 
similarly there are no compound noun phrases (``John and Fred went to the kitchen'').
Some of these seem easy to add to the simulation.
% but it should be easy to add in the future.
The hope is that adding these complexities will help evaluate models in a controlled way, within the simulated environment, which is hard to do with real data. 
Of course, this is not a substitute for real data which our models should be applied to as well, but does serve as a useful testbed.

\section{Word Sequence Training}\label{app:wordseq-training}

For segmenting an input word stream as generated in Appendix \ref{app:data-gen}
we use a segmenter of the form:
\[
   seg(c) = W_{seg}^\top U_{S} \Phi_{seg}(c) 
\]
where $ W_{seg}$ is a vector (effectively the parameters of a linear classifier in embedding space).
As we are already in the fully supervised setting, where for each question in the training
set we are given the answer and the supporting facts from the input stream, we can also use
that supervision for the segmenter as well. That is, for any known supporting fact, such
as ``Bill is in the Kitchen'' for the question ``Where is Bill?'' we wish the segmenter to
fire for such a statement, but not for unfinished statements such as ``Bill is in the''.
We can thus write our training criterion for segmentation as the minimization of:
\begin{equation}
  \sum\limits_{f \in {\cal F}} \max(0, \gamma - seg({f}))
 +  \sum\limits_{\bar{f} \in {\cal \bar{\cal F}}} \max(0, \gamma + seg(\bar{f}))
\end{equation}
where ${\cal F}$ are all known supporting segments in the labeled training set,
and $\bar{\cal F}$ are all other segments in the training set.

\section{Write Time Feature Training}\label{app:time-training}

The training procedure to take into account modeling write time is slightly different 
to that described in Section \ref{sec:basicmodel}.
Write time features are important so that the MemNN knows when each memory was written, 
and hence knows the ordering of statements that comprise a story or dialogue.
Note that this is different to time information described in the text of a statement,
 such as the tense
of a statement, or statements containing
 time expressions, e.g., ``He went to the office yesterday''. For such cases, write 
time features are not directly necessary, and they could (potentially)
be modeled directly from the text.

As was described in Section \ref{sec:writetime} we add three write time features
to the model and score triples using:
\begin{equation}
  s_{O_t}(x,y,y') = \Phi_x(x)^\top {U_{O_t}}^\top {U_{O_t}} \Big(\Phi_y(y) -\Phi_y(y') + \Phi_t(x,y,y')\Big).
\end{equation}
If $s_O(x,y,y') > 0$ the model prefers $y$ over
$y'$, and if $s_O(x,y,y')<0$ it prefers $y'$.  The $\mbox{argmax}$ of
eq. (\ref{eq:o1}) and (\ref{eq:o2}) are replaced by a loop over
memories $i=1,\dots,N$, keeping the winning memory ($y$ or $y'$) at each step,
and always comparing the current winner to the next memory
$\m_i$. That is, at inference time, for a $k=2$ model 
the $\arg\max$ functions of eq. (\ref{eq:o1}) and (\ref{eq:o2}) are replaced with
$o_1 = O_t(x, \m)$ and $o_2 = O_t([x, \m_{o_1}], \m)$ where 
$O_t$ is defined in Algorithm \ref{alg:o1} below.

\begin{algorithm}
 \caption{$O_t$ replacement to $\arg\max$ when using write time features
\label{alg:o1}
}
\begin{algorithmic}
\Function{$O_t$}{$q, \m$}
   \State{$t \gets 1$ }
   \For{$i=2,\dots,N$}
    \If{$s_{O_t}(q, \m_i, \m_t) > 0$}
    \State{$t \gets i$ }
    \EndIf
   \EndFor
   \State{\textbf{return} $t$}
\EndFunction
\end{algorithmic}
\end{algorithm}

$\Phi_t(x,y,y')$ uses three new features which take on the value 0 or
1: whether $x$ is older than $y$, $x$ is older than $y'$, and $y$
older than $y'$. When finding the second supporting memory (computing $O_t([x, \m_{o_1}], \m)$)
we encode whether $\m_{o_1}$ is older than $y$, $\m_{o_1}$ is older than $y'$,
and $y$ older than $y'$ to capture the relative age of the first supporting memory
w.r.t. the second one in the first two features.
 Note that when finding the first supporting memory (i.e., for $O_t(x, \m)$)
the first two features are useless as $x$ is the last thing in the memory and  hence $y$ and
$y'$ are always older.

To train our model with write time features 
we need to replace the hinge loss in eqs. (\ref{eq:train})-(\ref{eq:train2})
with a loss that matches Algorithm \ref{alg:o1}.
To do this, we instead minimize:
\begin{eqnarray*}
&  \sum\limits_{\bar{f} \neq \m_{o_1}}  \max(0, \gamma - s_{O_t}(x, \m_{o_1}, \bar{f})) + 
   \sum\limits_{\bar{f} \neq \m_{o_1}}  \max(0, \gamma + s_{O_t}(x, \bar{f}, \m_{o_1})) + \\
&  \sum\limits_{\bar{f'} \neq \m_{o_2}}  \max(0, \gamma - s_{O_t}([x, \m_{o_1}], \m_{o_2}, \bar{f'})) + 
   \sum\limits_{\bar{f'} \neq \m_{o_2}}  \max(0, \gamma + s_{O_t}([x, \m_{o_1}], \bar{f'}, \m_{o_2})  + \\
&   \sum\limits_{\bar{r} \neq r} 
      \max(0, \gamma - s_R([x, \m_{o_1}, \m_{o_2}], r) + s_R([x, \m_{o_1}, \m_{o_2}], \bar{r}]))
\end{eqnarray*}
The last term is the same as in eq. (\ref{eq:train3}) and is for the final ranking of words to return a response, which remains unchanged (as usual, this can also be replaced by an RNN for a more sophisticated model).
Terms 1-4 replace  eqs. (\ref{eq:train})-(\ref{eq:train2}) by considering triples directly.
For both $\m_{o_1}$ and $\m_{o_2}$ we need to have two terms considering them as the second or third
argument to $S_{O_t}$ as they may appear on either side during inference (via Algorithm \ref{alg:o1}). As before, 
at every step of SGD we sample $\bar{f}, \bar{f'}, \bar{r}$ rather than compute the
whole sum for each training example. 

\section{Word-sequence Learning Curve Experiments}\label{app:learning-curves}

We computed the test accuracy of MemNNs $k=2$ (+ time) for varying amounts of training
data: 100, 500, 1000 and 3000 training questions.
The results are given in Table \ref{table:wordlevel-learningcurve}.
These results can be compared with RNNs and LSTMs on the full data (3000 examples) by
comparing  with Figure \ref{table:simQA}.
For example, on the difficulty 5 actor and actor + object tasks MemNNs outperform LSTMs
even using 30 times less training examples.

\begin{table}[h!]
\caption{Test accuracy of MemNNs $k=2$ (+time) on the word-sequence simulation QA task for differing numbers of training examples (number of questions).
\label{table:wordlevel-learningcurve}
}
\begin{center}
\begin{tabular}{|l||c|c||c|c|}
\hline
&    \multicolumn{2}{c}{Difficulty 1} &  \multicolumn{2}{|c|}{Difficulty 5}  \\
\hline
Num. training       &   actor &  actor       &  actor      & actor \\
          questions &          &  + object   &             &  + object \\
\hline
%100    &  -\% & 45.7\% &  -\% & 33.4\% \\
100    &  73.8\%   & 64.9\% &  74.4\%    & 49.8\% \\
500    &  99.9\%   & 99.2\% &  99.8\%    & 95.1\% \\ 
1000   &  99.9\%   & 100\%  &  100\%    & 98.4\% \\ 
3000   &  100\%    & 100\% &  100\% & 99.9\% \\
\hline
\end{tabular}
\end{center}
\end{table}

\section{Sentence-level Experiments}\label{app:sentence-level}

We conducted experiments where input was at the sentence-level,
that is the data was already presegemented into statements and questions as input to the 
MemNN (as opposed to being input as a stream of words).
Results comparing RNNs with MemNNs are given in Table \ref{table:wordlevel-simQA}.
The conclusions are similar to those at the word level from Section 
\ref{sec:simulation}.
That is, MemNNs outperform RNNs, and that inference that finds $k=2$ supporting
statements and time features are necessary for the actor w/o before + object task.

\begin{table}[h!]
\caption{Test accuracy on the sentence-level simulation QA task.
\label{table:wordlevel-simQA}
}
\begin{center}
\begin{tabular}{|l||c|c||c|c|}
\hline
        &    \multicolumn{2}{c}{Difficulty 1} &  \multicolumn{2}{|c|}{Difficulty 5}  \\
\hline
       &   actor &  actor w/o before &  actor      & actor w/o before \\
Method &    w/o before &  + object   &  w/o before &  + object \\
\hline
% Difficulty 5:
RNN                    &  100\% & 58\%   &  29\% & 17\% \\
MemNN $k=1$             &  90\% & 9\%  &  46\% & 21\% \\
MemNN $k=1$ {(+time)}  &  100\% & 73\%    &  100\% & 73\% \\
MemNN $k=2$ {(+time)}   &  100\% & 99.95\%  &  100\% & 99.4\% \\
% Difficulty 1:
%RNN                   &  0\% & 42\%   \\
%MemNN $k=1$           &  10\% & 81\% \\
%MemNN $k=1$ {(+time)} &  0\% & 27\% \\
%MemNN $k=2$ {(+time)} &  0\% & 0.05\% \\
\hline
\end{tabular}
\end{center}
\end{table}

\section{Multi-word Answer Setting Experiments}\label{app:multiword}

We conducted experiments for the simulation data in the case 
where the answers are sentences (see Appendix \ref{app:data-gen}
 and Figure \ref{fig:rnn-story}). As the single word answer model can no longer be used,
we simply compare MemNNs using either RNNs or LSTMs for the response module $R$.
As baselines we can still use RNNs and LSTMs in the standard setting of being 
fed words only including the statements and the question as a word stream.
In contrast, the MemNN RNN and LSTMs are effectively fed the output of the 
$O$ module (see Section \ref{sec:basicmodel}). In these experiments
we only consider the difficulty 5 actor+object setting  in the case of MemNNs with 
$k=2$ iterations (eq. (\ref{eq:o2})), which means the module $R$ is fed
the features $[x, \m_{o_1}, \m_{o_2}]$  after the modules $I$, $G$ and $O$ have run.

The sentence generation is performed on the test data, and the evaluation we chose is as follows.
A correct generation has to contain the correct location answer, and can optionally contain 
the subject or a correct pronoun referring to it. For example the question ``Where is Bill?''
allows the correct answers ``Kitchen'', ``In the kitchen'', ``Bill is in the kitchen'',
``He is in the kitchen'' and ``I think Bill is in the kitchen''.
However incorrect answers contain an incorrect location or subject reference, for example
``Joe is in the kitchen'', ``It is in the kitchen'' or ``Bill is in the bathroom I believe''.
We can then measure the percentage of text examples that are correct using this metric.

The numerical results are given in Table \ref{table:mword-simQA}, and example output is given
in Figure \ref{fig:rnn-story}.
The results indicate that MemNNs with LSTMs perform quite strongly, 
outperforming MemNNs using RNNs. However, both MemNN variant outperform both RNNs and LSTMs
by some distance.

\begin{table}[h]
\caption{Test accuracy on the multi-word answer simulation QA task.
We compare conventional RNN and LSTMs with  MemNNs using an RNN or LSTM module $R$ ~(i.e., where $R$ is fed features $[x, \m_{o_1}, \m_{o_2}]$ after the modules $I$, $G$ and $O$ have run).
\label{table:mword-simQA}
}
\begin{center}
\begin{tabular}{|l|c|c|}
\hline
Model   & MemNN: IGO features $[x, \m_{o_1}, \m_{o_2}]$  &   Word features \\
\hline
RNN    & 68.83\%  & 13.97\% \\
LSTM   & 90.98\%  & 14.01\% \\
\hline
\end{tabular}
\end{center}
\end{table}

\end{document}